\documentclass[]{bytedance}
\usepackage[toc,page,header]{appendix}

\usepackage{minitoc}
\usepackage{amsfonts}
\usepackage{amssymb}
\usepackage{tabularx}
\usepackage{listings}
\usepackage{xcolor}
\usepackage{cancel}

\usepackage{multirow}
\usepackage{tabulary,multirow,xspace}
\usepackage{fixmath,mathtools,nicefrac,mmstyle}
\usepackage{subcaption}
\usepackage{caption}
\usepackage{wrapfig} 
\usepackage[misc]{ifsym} 
\usepackage{colortbl}
\usepackage{xcolor}
\usepackage{wrapfig}
\usepackage{multicol}
\usepackage[most]{tcolorbox}
\usepackage{pifont}

\definecolor{codegreen}{rgb}{0,0.6,0}
\definecolor{codegray}{rgb}{0.5,0.5,0.5}
\definecolor{codepurple}{rgb}{0.58,0,0.82}
\definecolor{backcolour}{rgb}{0.95,0.95,0.92}
\definecolor{boxblue}{RGB}{57,89,163}
\definecolor{boxbluebg}{RGB}{230,237,250} 

\lstdefinestyle{mystyle}{
    backgroundcolor=\color{backcolour},   
    commentstyle=\color{codegreen},
    keywordstyle=\color{magenta},
    numberstyle=\tiny\color{codegray},
    stringstyle=\color{codepurple},
    basicstyle=\ttfamily\footnotesize,
    breakatwhitespace=false,         
    breaklines=true,                 
    captionpos=b,                    
    keepspaces=true,                 
    numbers=none,                    
    numbersep=5pt,                  
    showspaces=false,                
    showstringspaces=false,
    showtabs=false,                  
    tabsize=2
}
\definecolor{mygray1}{gray}{.95}
\definecolor{mygray2}{gray}{.9}
\definecolor{mygray3}{gray}{.95}
\usepackage{pifont}

\newcommand{\myparagraph}[1]{{\noindent\bf #1}}

\newlength\savewidth
\newcolumntype{x}[1]{>{\centering\arraybackslash}p{#1pt}}

\newcommand{\app}{\raise.17ex\hbox{$\scriptstyle\sim$}}

\usepackage{xcolor}
\usepackage{graphicx}
\usepackage{amssymb}
\usepackage{pifont}
\usepackage{floatrow}
\usepackage{amsmath} 
\usepackage{float}
\usepackage{wrapfig}
\usepackage{multirow}
\usepackage{tcolorbox}
\usepackage{listings}
\usepackage{listings}
\usepackage{float}

\definecolor{commentgreen}{rgb}{0.1, 0.4, 0.1}
\definecolor{keywordblue}{rgb}{0.1, 0.1, 0.7}
\definecolor{stringred}{rgb}{0.7, 0.1, 0.1}

\lstdefinestyle{mystyle}{
    commentstyle=\color{commentgreen},
    keywordstyle=\color{keywordblue},   
    stringstyle=\color{stringred},
    basicstyle=\ttfamily\scriptsize, 
    breaklines=true,
    keepspaces=true,
    showstringspaces=false,
    frame=none,                     
    language=Python, 
}

\newcommand{\name}{OmniHuman-1.5}
\title{\name{}: Instilling an Active Mind in Avatars via Cognitive Simulation}

\author{
\centerline{
Jianwen Jiang $^{*\dagger}$\quad 
    Weihong Zeng $^*$ \quad  
    Zerong Zheng $^*$ \quad 
    Jiaqi Yang $^*$ \quad
} 
\centerline{
    Chao Liang $^*$ \quad
    Wang Liao $^*$ \quad
    Han Liang $^*$ \quad 
    Yuan Zhang  \quad 
    Mingyuan Gao \quad 
}
}

\affiliation[]{Intelligent Creation Lab, ByteDance}

\footnotetext{$^*$Equal contributions}
\footnotetext{$^\dagger$Project lead and corresponding author: jianwen.alan@gmail.com}

\abstract{

Existing video avatar models can produce fluid human animations, yet they struggle to move beyond mere physical likeness to capture a character's authentic essence. Their motions typically synchronize with low-level cues like audio rhythm, lacking a deeper semantic understanding of emotion, intent, or context. To bridge this gap, \textbf{we propose a framework designed to generate character animations that are not only physically plausible but also semantically coherent and expressive.} Our model, \textbf{OmniHuman-1.5},  is built upon two key technical contributions. First, we leverage Multimodal Large Language Models to synthesize a structured textual representation of conditions that provides high-level semantic guidance. This guidance steers our motion generator beyond simplistic rhythmic synchronization, enabling the production of actions that are contextually and emotionally resonant. Second, to ensure the effective fusion of these multimodal inputs and mitigate inter-modality conflicts, we introduce a specialized Multimodal DiT architecture with a novel Pseudo Last Frame design. The synergy of these components allows our model to accurately interpret the joint semantics of audio, images, and text, thereby generating motions that are deeply coherent with the character, scene, and linguistic content. 
Extensive experiments demonstrate that our model achieves leading performance across a comprehensive set of metrics, including lip-sync accuracy, video quality, motion naturalness and semantic consistency with textual prompts. Furthermore, our approach shows remarkable extensibility to complex scenarios, such as those involving multi-person and non-human subjects.

}

\date{\today}

\checkdata[Project Page]{\url{https://omnihuman-lab.github.io/v1_5}}

\begin{document}
\maketitle

\begin{figure}[ht]
  \centering
  \includegraphics[width=\linewidth]{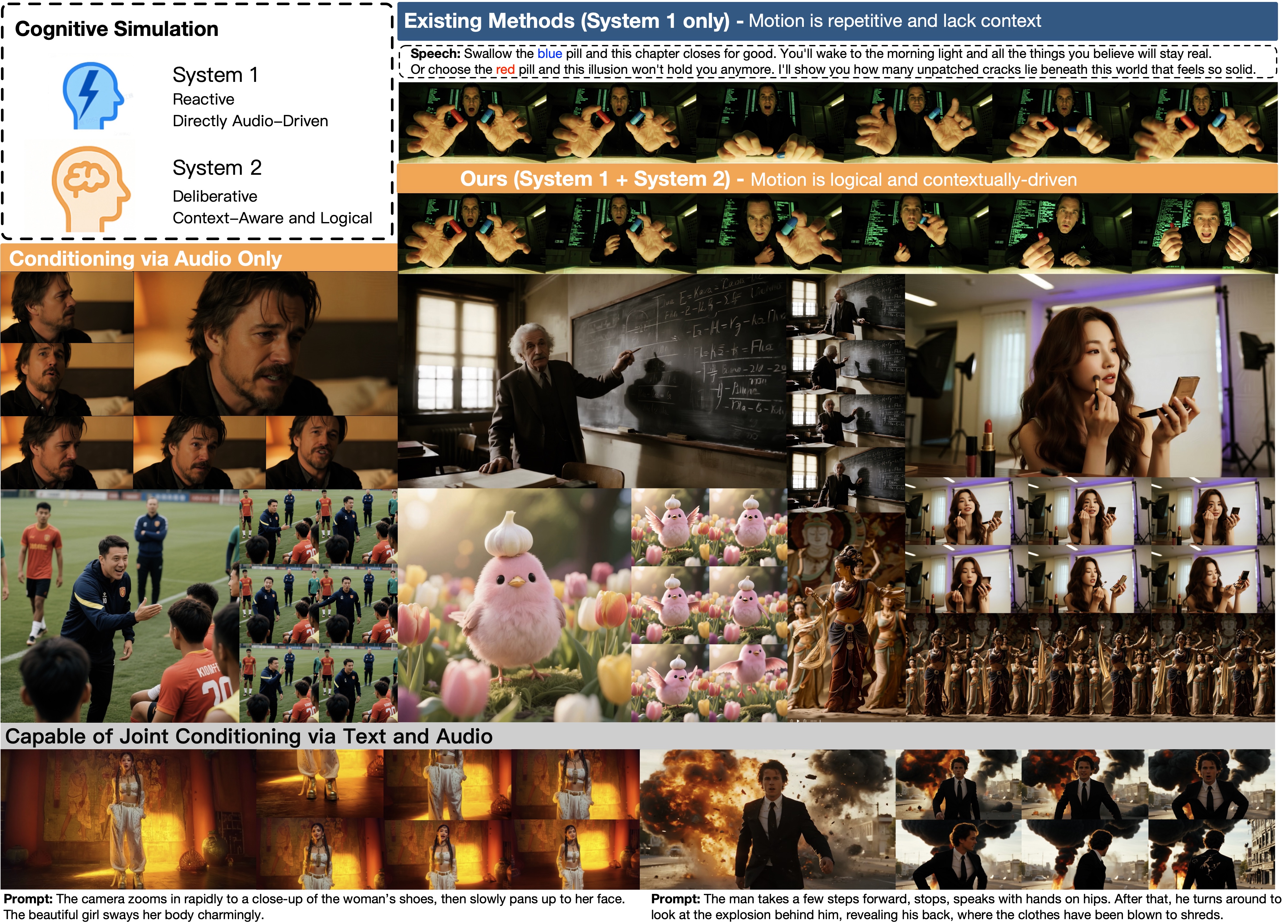}
  \caption{\textbf{Simulating a Mind for Avatars.} We model avatar behavior by drawing on dual-system theory, which distinguishes between reactive System 1 and deliberative System 2 cognition. Top Left: Our framework combines System 1 actions (e.g., lip-sync, idle motions) with System 2 reasoning (e.g., logical gestures). Top Right: Conventional methods, analogous to System 1, excel at lip-sync but often produce repetitive, non-contextual motions. Bottom: In contrast, our method simulates both systems, generating diverse and naturally coherent behaviors that are semantically aligned with the provided audio and text.}
  \label{fig:teaser}
\end{figure}

\section{Introduction}\label{sec:intro}

\begin{quote}
"System 1 operates automatically and quickly, with little or no effort and no sense of voluntary control. System 2 allocates attention to the effortful mental activities that demand it, including complex computations."
\hfill\null 
\par\nopagebreak
\hfill --- Daniel Kahneman, \textit{Thinking, Fast and Slow}
\end{quote}

The field of video avatars~\cite{he2023gaia,tian2024emo,xu2024hallo,wang2024vexpress,chen2024echomimic,xu2024vasa,stypulkowski2024diffused,jiang2025loopy,lin2025cyberhost,gan2025omniavatar,kong2025multitalk,wang2025fantasytalking,hu2024animate,lin2025omnihuman,qiu2025skyreels} aims to construct models capable of synthesizing realistic videos of characters from human-centric driving signals. The ultimate goal can be seen as creating lifelike avatars that are virtually indistinguishable from humans, capable of demonstrating both reasoned action and authentic emotion. This domain has witnessed rapid advancements over the past few years, with model capabilities evolving from early lip movement synthesis~\cite{jiang2024mobileportrait,zhang2023sadtalker,facev2v,zhao2022tps,fomm,mraa} and portrait animation~\cite{jiang2025loopy,tian2024emo,fada,xu2024vasa,xu2024hallo,hallo3} to more recent half-body~\cite{lin2025cyberhost,emo2} and full-body generation~\cite{lin2025omnihuman}. As the scope of controllable generation expands, these models are increasingly expected to simulate a broader spectrum of human behaviors, moving beyond mere physical likeness to capture a character's authentic essence, as illustrated in Figure~\ref{fig:teaser}.

Recently, a series of audio-driven methods~\cite{lin2025omnihuman,kong2025multitalk,gan2025omniavatar,wang2025fantasytalking, qiu2025skyreels} based on Diffusion Transformers (DiT)~\cite{dit,esser2024sd3,seawead2025seaweed,kong2024hunyuanvideo,wan2025} have emerged, enabling the generation of human motion videos that synchronize with audio in specific contexts. These approaches typically follow a similar paradigm: they first pre-train a foundational video generation model~\cite{wan2025,seawead2025seaweed} and then introduce conditional audio inputs to achieve animation. While this yields reasonable results, a closer inspection reveals that these models often only capture the direct and simplistic correlations between the audio signal and the resulting motion. Consequently, they tend to generate only synchronized lip movements and simple, repetitive accompanying gestures. When judged on their potential to function as convincing avatars, these outputs still exhibit a significant gap in naturalness and plausibility when compared to authentic human behavior.

To understand the source of this gap, it is instructive to consider a prominent theory in human cognition, which posits that behavior is governed by two distinct systems~\cite{kahneman2011thinking,kahneman2013prospect}, often termed System 1 and System 2. System 1 is described as fast, unconscious, and reactive. This operational mode is analogous to that of current video avatar models, which excel at mapping input signals like audio directly to corresponding lip movements and simple gestures. In contrast, System 2 is characterized as deliberative, analytical, and effortful, involving reasoning about goals and context to deduce a course of action. This latter capability, the hallmark of intelligent behavior, remains challenging for existing methods to emulate robustly. Inspired by this dual-process model, we identify a clear path forward: leveraging the powerful reasoning capabilities~\cite{wei2022chain, wangself, yao2023tree, schick2023toolformer, park2023generative, yuan2024mora, wu2025automated} of modern Multimodal Large Language Models (MLLMs)~\cite{team2023gemini,GPT4o} to explicitly simulate the deliberative processes of System 2. We therefore propose a novel framework that integrates the principles of both systems, using MLLMs to simulate the goal-oriented aspects of System 2 while preserving the reactive capabilities analogous to System 1.

However, integrating the textual guidance from an MLLM into an avatar generation framework is non-trivial. Leveraging Chain-of-Thought (CoT)~\cite{wei2022chain} prompting, MLLMs articulate their reasoning process as explicit text, a natural medium for expressing logic. Consequently, introducing an MLLM to enhance the high-level semantic coherence of motion inevitably strengthens the role of text as a conditional input. This creates a potential conflict of modalities within existing avatar models, where the audio signal already governs lip-sync and rhythmic motion, and the reference image constrains the character's appearance and potential range of motion. These distinct inputs are all intricately linked to the final generated motion. Therefore, we argue that a novel framework must be designed to mitigate modal conflicts while effectively leveraging the interdependencies among these diverse conditions. Developing such a framework is the key to simultaneously simulating both ``System 1'' and ``System 2''.

Based on the preceding analysis, we propose a video avatar framework built upon a Multimodal Diffusion Transformer that features two key designs. First, to ensure a comprehensive simulation of both System 1 and System 2, we introduce an agentic system powered by MLLMs. These agents reason over multimodal inputs (text, reference image, audio) to generate high-level semantic context, providing a long-term, logically coherent signal (for ``System 2'') that complements the short-term, reactive audio signals (for ``System 1'').
Second, to effectively fuse these input conditions while mitigating modality interference, we employ dedicated Multimodal Branches for audio and text feature extraction, along with a Multimodal Attention mechanism for joint modeling of audio, text and video. This design iteratively updates and aligns audio and text features into a common semantic space. Furthermore, we introduce a distinct identity preservation strategy that avoids conditioning on the reference image during training. Instead, we guide the model by probabilistically conditioning on start/end frames during training and treating the reference image as a pseudo last frame at inference, which prevents the static image from interfering with dynamic, content-driven motion.
As a result of these designs, our model generates vivid and lifelike human motion from audio and a single image. The resulting videos exhibit remarkable contextual and semantic coherence, effectively simulating both the reactive and deliberative aspects of human behavior.

We summarize our main contributions as follows:
\begin{itemize}
    \item \textbf{A New Perspective on Avatar Modeling:} We introduce a new perspective for analyzing video avatars, framing the problem through the cognitive science lens of System 1 and System 2 thinking. Observing that current models primarily simulate System 1, we are the first to propose a holistic approach that models both.
    \item \textbf{A Framework for Dual-System Simulation:} To implement this vision, we propose a novel framework featuring two core components. First, MLLM-based agents generate deliberative guidance (``System 2''). Second, a specialized MMDiT architecture, equipped with a symmetric audio branch and our pseudo-last-frame strategy, synergistically fuses this guidance with reactive signals (``System 1'') , thereby resolving critical modal conflicts.
    \item \textbf{Strong Empirical Performance and Generalization:} Our method not only achieves highly competitive results on standard benchmarks but is also significantly preferred in user studies for its contextual naturalness and plausibility. Its versatility is further demonstrated by its successful extension to complex multi-person and non-human scenarios.
\end{itemize}

\section{Related Work}\label{sec:related}

\myparagraph{Video Generation.} The field of video generation has seen rapid advancements, largely building upon the success of diffusion models in visual synthesis~\cite{ho2020denoising, song2020denoising}. Current approaches can be broadly categorized based on their underlying architecture.
The first category consists of methods based on pre-trained text-to-image U-Net models~\cite{esser2024scaling, chen2024pixart}. These approaches typically insert temporal modules, such as attention or convolution layers, into the frozen U-Net backbone and then fine-tune the model on video data~\cite{guo2023animatediff, wang2023modelscope}. This allows them to leverage powerful image priors, but their capabilities can be constrained by the original image-centric architecture.
The second category is represented by models that adopt a Diffusion Transformer (DiT) architecture~\cite{brooks2024video, yang2024cogvideox, zheng2024open, kong2024hunyuanvideo, ma2025step, polyak2024movie,chen2024gentron,menapace2024snap}. These methods treat video as a sequence of spatiotemporal patches, processing them in a unified manner with a Transformer. This approach has demonstrated superior scalability and flexibility, enabling the generation of high-resolution videos with variable durations and aspect ratios, especially when trained on massive datasets.
The third category explores the integration of Large Language Models (LLMs) with diffusion models. In the image generation domain, works such as~\cite{yu2023scaling, koh2023generating,pan2025transfer,wu2025qwen,wu2024next,shi2024lmfusion}have shown that LLMs can enhance compositional understanding and planning. For video generation, this area is still in a nascent stage but holds significant promise for improving the logical coherence and narrative structure of generated content.

\myparagraph{Video Avatar Model.} Video avatars aims to create realistic human videos from various driving signals, with methods typically categorized by their driving source and synthesis pipeline. One category is pose-driven animation, which generates video based on an explicit, externally provided motion sequence, such as skeletal poses~\cite{zhang2024mimicmotion, shao2024human4dit,chang2023magicpose,tu2024motionfollower,wang2024unianimate,karras2023dreampose,xu2024magicanimate}. As the motion generation step is largely bypassed in these methods, their focus shifts to achieving high-fidelity rendering. More central to this paper is the second key category: audio-driven animation. This task requires a model to first generate plausible human motion from an audio signal and subsequently synthesize the final video. Consequently, the model must handle the dual challenges of both motion generation and rendering. Within this category, a common approach is a two-stage pipeline, where a motion generation model first translates audio into an intermediate representation, such as 2D/3D keypoints or 3D mesh sequences~\cite{zhuang2024vlogger, meng2025echomimicv2,deng2025stereo,wei2024aniportrait} or discrete motion tokens~\cite{hogue2024diffted, tian2025emo2}. A subsequent rendering model then synthesizes the video conditioned on this motion. More recently, end-to-end approaches have emerged~\cite{lin2025cyberhost,lin2025omnihuman,wang2025interacthuman,liang2025alignhuman}, which directly generate video from audio in a single step, aiming for improved audio-motion synchronization.
Despite their architectural differences, these audio-driven methods predominantly treat motion generation as a direct mapping process, a fast, reactive function from audio features to motion output. This process does not explicitly model the high-level cognitive phase where humans' planning and reasoning ultimately determine the resulting physical actions. We believe that incorporating such a reasoning phase is essential for generating more plausible and intelligent human behaviors, and our work aims to provide a foundational exploration in this direction.

\myparagraph{Large Language Models on Cognitive Simulation.} Large Language Models (LLMs) have reshaped machine reasoning with a clear upward trajectory in cognitive capability. Foundational models like GPT-3~\cite{brown2020language} and ChatGPT~\cite{ouyang2022training, achiam2023gpt} laid the groundwork, while recent iterations such as GPT-4o~\cite{hurst2024gpt}, OpenAI o-series~\cite{jaech2024openai} and DeepSeek-R1~\cite{guo2025deepseek} enhance contextual and multi-modal reasoning, with prompting being key to unlocking this potential—Chain-of-Thought (CoT)~\cite{wei2022chain} revolutionized problem-solving by decomposing tasks into logical steps. Derivatives like Self-Consistency~\cite{wangself} using aggregating reasoning paths and Tree-of-Thought (ToT)~\cite{yao2023tree} with branching exploration further improved accuracy in arithmetic, logic, and commonsense tasks, confirming LLMs can simulate human-like deductive/inductive thinking via structured prompting. 
This reasoning capability has further empowered autonomous agents as their core engine. Classic practical frameworks like Toolformer~\cite{schick2023toolformer}, MetaGPT~\cite{hong2023metagpt}, and AutoGPT~\cite{zhang2023autogpt} with autonomous goal refinement enable agents to transcend pre-defined rules, exhibiting intent understanding and error correction. Voyager~\cite{wang2023voyager} leverages LLMs for open-world strategies and adaptive responses, outperforming rule-based agents. Furthermore, the renowned Generative Agents~\cite{park2023generative} enable LLMs to simulate daily human behaviors (e.g., social interaction, scheduling), marking a pivotal shift in human-like agent simulation. Recent advancements like OpenAI's Deep Research~\cite{openai2025deepresearch} and Monica's Manus~\cite{monica2025manus} further showcase LLM-driven proficiency in complex task planning and cross-domain execution.
Beyond agent systems, LLM reasoning also steers generative models across domains to solve the "controllable generation" challenge. For image editing, InstructPix2Pix~\cite{brooks2023instructpix2pix} relies on LLM-generated instructions to refine image modifications. MetaQueries~\cite{pan2025transfer} uses learnable queries to prompt vision-language models to guide image generation with enhanced semantic alignment. Besides, recent LLM-driven video generation agents~\cite{hu2024storyagent, yuan2024mora, li2024anim, wu2025automated} enable more controllable long video synthesis through collaborative agentic workflows. These works collectively position LLMs as universal planners that bridge high-level user intent and the execution logic of low-level generative models. Yet notably, the integration of LLM-driven reasoning and planning into more fine-grained intelligent human/avatar behavior generation remains an underexplored area.

\section{Approach}\label{sec:method}

\subsection{Overview}

\begin{figure}[t]
  \centering
  \includegraphics[width=\linewidth]{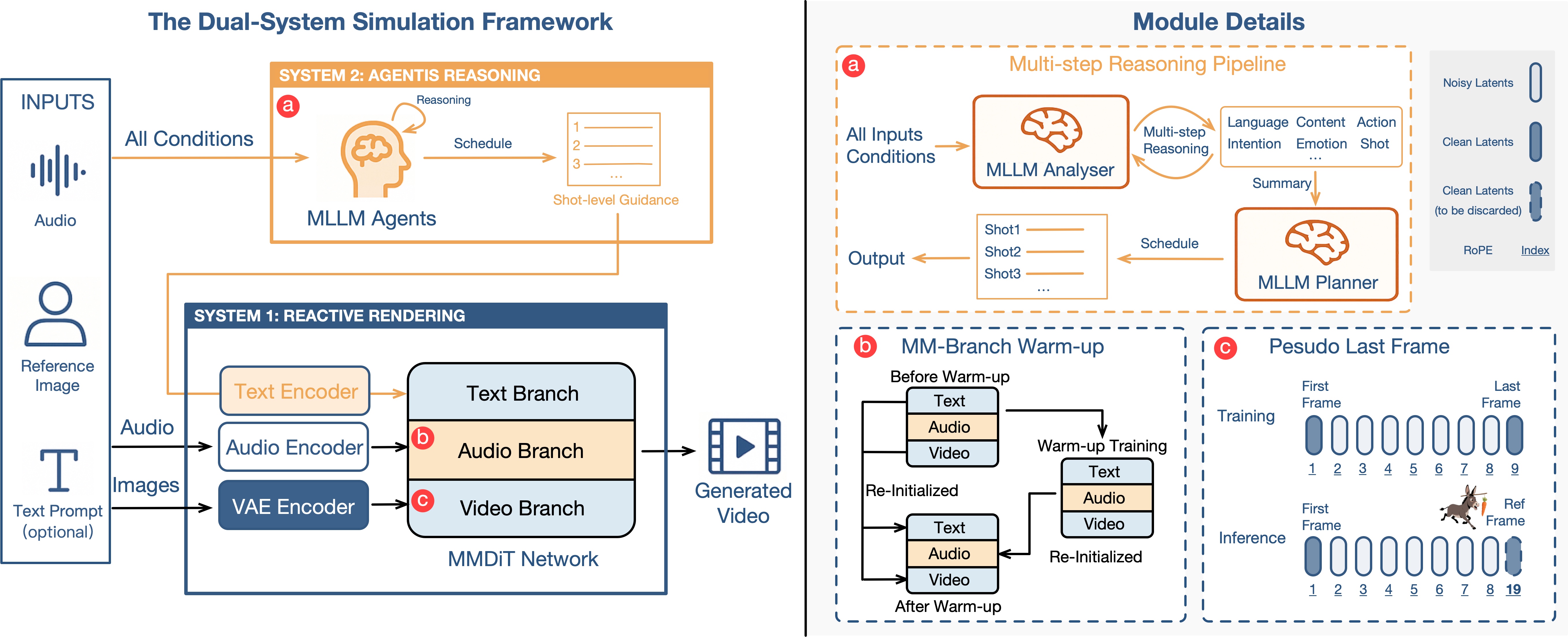}
  \caption{\textbf{The Dual-System Simulation Framework.} Our framework models avatar behavior by integrating a deliberative \textbf{System 2} for planning with a reactive \textbf{System 1} for rendering. \textbf{Left:} The overall pipeline. System 2 uses an MLLM Agent to reason over all inputs (audio, image, text) and generate a high-level "schedule". This schedule guides System 1's MMDiT network, which synthesizes the final video by fusing information within its dedicated text, audio, and video branches. \textbf{Right:} Key module details. (a) The reasoning pipeline consists of an MLLM Analyser and Planner that work together to create the schedule. (b,c) Our proposed \textbf{MM-Branch Warm-up} and \textbf{Pseudo Last Frame} methods mitigate multimodal conflicts during training.}
  \label{fig:framework}
\end{figure}

Our goal is to generate character animations that are both visually realistic and logically coherent with multimodal inputs. To achieve this, we introduce a framework designed to simulate both System 1 (reactive) and System 2 (deliberative) cognitive processes. Our model is built upon a Diffusion Transformer (DiT) backbone~\cite{seawead2025seaweed,gao2025seedance,esser2024sd3,dit}, which is first pre-trained~\cite{gao2025seedance} on general video generation tasks to acquire foundational video generation capabilities. We then transform this base model into a logical and expressive avatar through two critical designs, which are detailed in the following sections and illustrated in Figure~\ref{fig:framework}.

\par
Agentic Reasoning for Deliberative Control (Sec. 3.2): We first employ MLLM-based agents to reason about the input context and generate high-level semantic guidance. This step provides the deliberative control necessary to simulate ``System 2''. 

Multimodal Fusion for Reactive Rendering (Sec. 3.3): Next, our specialized MMDiT~\cite{esser2024sd3} architecture fuses this semantic guidance with reactive signals like audio  to simulate ``System 1''. To resolve modal conflicts, this architecture incorporates a pseudo-last-frame identity strategy, which prevents the static reference image from interfering with dynamic motion during training. 
\par
Beyond these core designs, our framework aligns with common practices~\cite{lin2025omnihuman,gan2025omniavatar,kong2025multitalk} for simplicity. To support long-form video synthesis, our framework can operate autoregressively, continuing generation by using the final frames of a previously generated clip as the initial frames for a new segment~\cite{stypulkowski2024diffused}. Our framework operates in the compact latent space of a pre-trained 3D VAE~\cite{yu20233DVAE} and is trained with a flow matching~\cite{liu2022reflow} objective. We omit further discussion of these standard components to focus on our main contributions.

\subsection{Agentic Reasoning for Deliberative Control}

\myparagraph{Module Inputs and Outputs.} To model the deliberative nature of ``System 2'', our Agentic Reasoning module reasons over the input conditions to generate high-level, logically coherent guidance. The inputs are the character's reference image and the corresponding audio clip, supplemented by an optional text prompt describing the desired character behavior. The module is designed to process these inputs to produce semantic conditions in two forms: Reasoning Text, the explicit chain-of-thought text from the agents used directly as a condition, and Reasoning Latents, intermediate MLLM features extracted to serve as an additional conditioning signal and integrated via a dedicated attention mechanism. Our primary approach utilizes the former, while we have also investigated the integration of the latter.

\myparagraph{Multi-Step Reasoning Pipeline.} As shown in the top-right portion of Figure~\ref{fig:framework}, this deliberative guidance is generated using two MLLMs that play distinct roles in a collaborative process. The first MLLM, the Analyzer, receives the reference image, a corresponding text caption (produced by an auxiliary model to ensure accurate image interpretation), the audio clip, and an optional user-provided text prompt. Guided by a chain-of-thought prompt, the Analyzer MLLM performs an iterative reasoning process on the inputs. It systematically infers the character's persona, language style, speech content, emotion, intent, and environmental context, consolidating these insights into a single structured representation, typically a JSON object. This context-rich information is then passed to the second MLLM, the Planner. The Planner receives the Analyzer's output, along with the original character image for visual context. Guided by a new instructional prompt, its task is to devise an action plan structured as a sequence of shots, where each shot defines the character's expressions and actions for a duration corresponding to a single generation pass of our diffusion model. This collaborative reasoning yields a comprehensive motion schedule that preserves a coherent character persona through consistent actions across the entire video.

\myparagraph{Reflective Re-planning.} To maintain logical coherence in long-form video generation, our agentic framework incorporates an optional ``reflection'' process. During autoregressive synthesis, the generation schedule is dynamically updated by re-evaluating the most recently generated output. This process mitigates a common challenge in diffusion-based synthesis, where subtle execution deviations can accumulate and degrade logical coherence over time, particularly in longer videos.
In practice, the Planner takes the last generated frames and the original reference image as new inputs to re-assess its plan. This reflective loop corrects for semantic drift and helps maintain the video's logical consistency.

 \myparagraph{Investigation of Latent Feature Conditioning.} We also investigated the aforementioned approach of utilizing latent features from a MLLM as a semantic conditioning signal. The approach directly utilized the audio tokens within the transformer of the Analyzer agent. The rationale was that cross-modal attention in the transformer layers would enrich these audio token representations with high-level semantics (e.g., inferred emotion and intent) while preserving their original temporal structure. Based on this consideration, we selected these ``reasoning-infused'' audio latents from the final transformer layer and concatenated them with the raw audio features. This combined signal then replaced the original audio input for the DiT network.
\par

The above designs enable our agent to formulate a global, coherent plan for the entire scene. Unlike purely reactive methods that merely emulate ``System 1'', our approach further integrates deliberative reasoning from ``System 2'' to provide thoughtful, top-down guidance.

\subsection{Reactive Rendering via Multimodal Diffusion}

In this subsection, we describe how our diffusion model synthesizes the final video. It synergistically combines the high-level reasoning from the agents (primarily represented as text) with the low-level, reactive signals of audio inputs (primarily represented as audio features).

\begin{wrapfigure}{r}{0.5\textwidth}
    \centering
    \vspace{-12pt}
    \includegraphics[width=\textwidth]{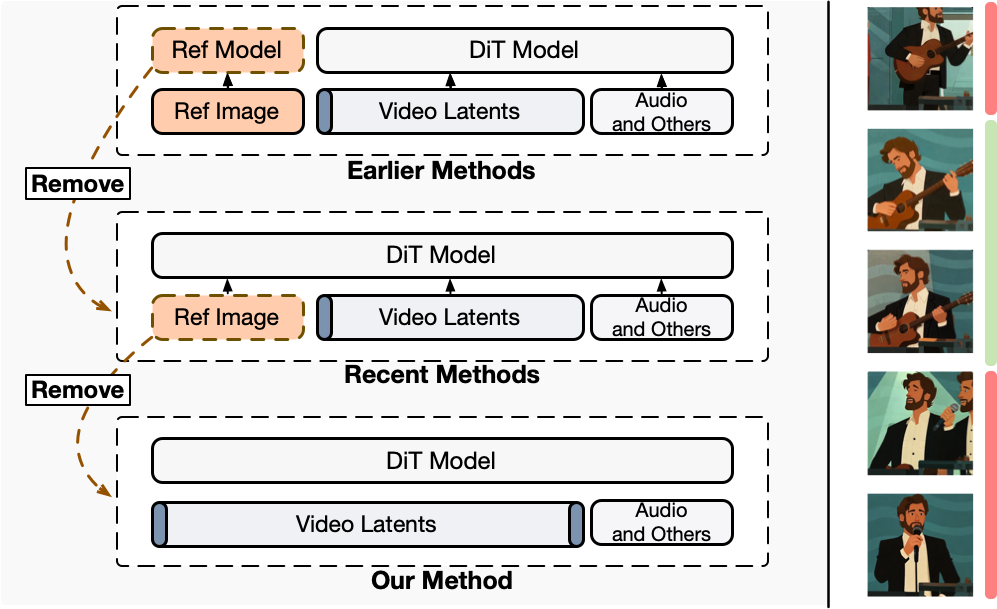}
    \vspace{-18pt}
      \caption{\textbf{Rationale for the Pseudo Last Frame.} \textbf{Left:} Reference-conditioning has trended toward simplification. \textbf{Right:} The dilemma of reference image sampling. Sampling from \textcolor{green!50!black}{\textbf{within}} the target video segment ensures high relevance but restricts motion diversity. Conversely, sampling from \textcolor{magenta}{\textbf{outside}} the segment, a scenario that becomes more frequent as datasets grow larger and more dynamic, leads to a drop in content relevance, causing inconsistencies that undermine the reference's purpose.}
  \label{fig:plf}
\end{wrapfigure}

\myparagraph{Rethinking Reference Conditioning.} Before detailing our driving condition modeling, we must first analyze a critical input in video avatar models: the conditioning image, which serves two distinct purposes. The first is providing initial frames for autoregressive continuity, a standard practice we adopt by concatenating ground-truth (GT) frames. The second, more problematic purpose, is using a reference image for identity preservation. While early works used dedicated networks~\cite{aa,tian2024emo,zhu2023tryondiffusion} and recent methods reuse model parameters~\cite{lin2025omnihuman,kong2025multitalk}, both approaches inject a reference image sampled from the training video. As illustrated in Figure~\ref{fig:plf}, we argue that this creates a critical artifact: the model learns a spurious correlation that the reference image should appear literally within the generated sequence, severely restricting motion dynamics and conflicting with the audio and text driving signals. While this artifact can be partially mitigated by probabilistically sampling reference images from outside the training video clip, this approach may introduce new problems, causing the model to learn that the generated output should exhibit significant variation from the reference image. 

\par
The root cause of this issue is that the reference image is an artificial construct, not a condition native to the video data itself. It is for this reason that our solution is to discard it entirely during training and, in its place, introduce a novel guidance mechanism. As shown in the bottom right of Figure~\ref{fig:framework}, during training, we probabilistically condition the model on both the GT first and last frames of the video clip, as these are both native signals. During inference, we repurpose this mechanism by placing the user's reference image in the last frame's position, creating a pseudo last frame. Crucially, we shift its positional encoding (e.g., RoPE~\cite{rope}) to maintain a fixed temporal distance from the generated content. This pseudo-frame, which is dropped after rendering, functions as a ``carrot on a stick'': it guides the model toward the reference identity without ever forcing it to replicate the image. As our experiments show, this approach eliminates training artifacts and mitigates autoregressive error, achieving a superior trade-off between motion dynamics and stability.

\myparagraph{Symmetric Fusion and Warm-Up.} With all training conditions now native and compatible, we address the challenge of joint modeling. We adopt an MMDiT backbone but depart from prior work in our approach to audio conditioning. Instead of injecting audio features via additional cross-attention layers, we introduce a dedicated audio branch, architecturally symmetric to the video and text branches. All three modalities are then fused at each layer through a \textbf{shared} multi-head self-attention mechanism. This symmetric design offers two key advantages. First, it allows audio features to be iteratively refined alongside video and text, ensuring deep semantic alignment. Second, it enables true joint modeling, as tokens from all three modalities mutually attend to one another, facilitating a more effective mapping to a shared semantic space. While this new branch adds parameters, the computational overhead is negligible due to the low ratio of audio to video tokens.
\par
This symmetric architecture, while beneficial, introduces a training challenge. Naively training the entire model jointly leads to modality conflict, where the model learns to over-rely on the temporally dense audio signal for all predictions, thereby ignoring or washing out the more abstract guidance from the text branch. Freezing the pre-trained branches is also suboptimal, as it causes the audio branch to overfit and erroneously learn non-audio-related attributes like lighting and camera motion. To resolve this, we propose a two-stage warm-up strategy. In Stage 1, we train the full three-branch model jointly, forcing the model to learn an optimal division of labor: the text and video branches handle high-level semantics, compelling the audio branch to specialize in its core competencies (e.g., lip sync, speech mannerisms). For Stage 2, we construct the final model. The text and video branches are initialized with their original pre-trained weights, while the audio branch is initialized using the warmed-up weights obtained from Stage 1. This model is then fine-tuned. This strategy ensures each branch begins with its own strong, specialized prior, mitigating modality conflict and allowing each input to retain its distinct conditioning power. 
\par
Ultimately, the proposed architecture executes the deliberative plan. By redesigning the reference conditioning and equipping the model with an audio conditioning branch, our rendering process faithfully translates the high-level guidance from System 2 while maintaining the reactive fidelity of System 1.
\section{Experiments}\label{sec:exp}

\subsection{Experimental Setup}

\myparagraph{Implementation Details.}
Our model is based on the MMDiT architecture and was pre-trained on a large-scale dataset of text-video/image pairs. For most experiments, the model generates 120-frame clips at 24~fps with a 480p resolution (short side). A separate super-resolution model of the same architecture is used to upscale outputs to 720p or 1080p. Longer videos are generated autoregressively. We used the AdamW optimizer with a learning rate of 5e-5, a global batch size of 256, and gradient clipping at a norm of 1.0. The training was conducted on 256 compute nodes and consisted of three stages: a 3-day audio branch warm-up, a 7-day main training phase, and a 1-day fine-tuning phase on high-quality data.

\myparagraph{Training Data.}
Our training set consists of 15,000 hours of filtered video data. Following prior work, we used a lip-sync model to identify and discard the audio from videos with poor lip-audio correlation. These samples, comprising 70\% of the data, were used with audio-dropout during training. For the final fine-tuning stage, we ranked our training data by quality metrics and selected the top 100 hours.

\myparagraph{Evaluation Datasets.}
To rigorously evaluate our model, we noted that current DiT-based methods already perform well in standard human speaking scenarios. To test the true generalization limits of our approach, we therefore constructed two novel and highly challenging test sets. Our first custom benchmark is a diverse single-subject set of 150 cases, including real-world human portraits, AIGC figures, anime characters and animals. Each image was manually paired by experts with a corresponding audio track, such as speech, singing or theatrical performances, to create a demanding generalization test. To assess performance in more complex scenes, we also built a multi-subject set of 57 cases, featuring the same visual diversity and expert-paired audio for multi-character interactions. Furthermore, to evaluate our model's text-conditioning, experts wrote descriptive prompts for all 150 single-subject cases, allowing us to measure adherence to textual guidance. Finally, for fair comparison with prior work, we adopted their experimental settings, using 100 videos from CelebV-HQ~\cite{zhu2022celebv} for the talking-head task and the CyberHost~\cite{lin2025cyberhost} test set (269 videos, 119 identities) for evaluating performance in full-body scenarios.

\myparagraph{Evaluation Metrics.}
To comprehensively assess our method, we employ a multi-faceted evaluation protocol that includes both objective and subjective metrics.
For objective evaluation, we measure generation quality using the Fr\'echet Inception Distance (FID)~\cite{heusel2017gans} and Fr\'echet Video Distance (FVD)~\cite{unterthiner2019fvd}, alongside no-reference Image Quality (IQA) and Aesthetics (ASE) scores~\cite{wu2023q}. We also evaluate audio-visual synchronization with Sync-C~\cite{syncnet}, hand quality using Hand Keypoint Confidence (HKC) and Hand Keypoint Variance (HKV)~\cite{lin2025cyberhost}.
\par
However, as these objective metrics often fail to capture higher-level semantic qualities and overall perceptual realism, we also conducted a comprehensive subjective user study with 40 participants. This study involved two main protocols. The first was a \textbf{pairwise comparison}, where participants viewed two videos from different methods in a randomized order. This comparison was twofold: from a positive perspective, users selected the video with the best overall quality, from which we calculated the Good/Same/Bad (GSB) score, defined as $(\text{Wins} - \text{Loses}) / (\text{Wins} + \text{Loses} + \text{Ties})$; from a critical perspective, they identified specific flaws: Lip-sync Inconsistency (LSI), Motion Unnaturalness (MU), and Image Distortion (ID), allowing us to compute a defect rate for each. The second protocol was a \textbf{best-choice selection task}, where participants selected the single best video from all competing methods. This yielded a Top-1 selection rate, providing a direct measure of overall appeal.

\subsection{Ablation Studies}
In this section, we conduct a series of ablation studies to rigorously validate the contributions of our proposed components. The experiments are performed on a custom, single-subject test set of 150 video clips. Our analysis systematically isolates the impact of two key elements: (1) the agentic reasoning module and (2) the proposed conditioning architecture within our diffusion model. For a comprehensive assessment, we employ both quantitative and subjective evaluations. The quantitative metrics provide an objective measure of performance, while the subsequent user studies assess perceptual quality in terms of lip-sync consistency, motion naturalness, image quality, and overall user preference.

\myparagraph{The Effectiveness of Agentic Reasoning.}
Here, we analyze the contribution of the Agentic Reasoning module by dissecting its intermediate steps. We conduct experiments by first removing the multi-step reasoning process, then ablating the entire Analyzer, and finally, removing the reasoning module altogether, which results in a ``System 1 only'' model. As shown in Table~\ref{tab:full_ablation}, standard metrics for image quality (IQA) and lip-sync (Sync-C) show only minor variations across these ablations. This is expected, as these metrics primarily evaluate low-level fidelity, which remains high in all diffusion-based variants. However, they are not designed to measure higher-level semantic qualities like logical coherence. A more telling objective trend emerges with the HKV metric, which progressively decreases as reasoning capabilities are weakened, indicating that the generated animations become more static and less expressive.
\par
To truly evaluate the core contributions of our reasoning module to these semantic qualities, we therefore turn to subjective evaluation. The results, presented in (a) of Table~\ref{tab:subjective_ablation_full}, offer a direct comparison between the models with and without the agentic reasoning module. The overall user preference (GSB score) immediately reveals a substantial advantage for our full model. More specifically, the introduction of reasoning leads to a significant reduction in perceived motion unnaturalness (MU), with over a 20\% improvement in pairwise comparisons. Furthermore, it maintains or slightly improves lip-sync consistency and image quality, as reflected by the LSI and ID metrics. These findings support the effectiveness of our Agentic Reasoning module, particularly in its ability to enhance the plausibility and semantic motion naturalness of the generated animations, which are qualities not fully captured by objective metrics.

\begin{table}[t!]
    \centering
    \caption{\textbf{Ablation studies on our proposed framework.} }
    \label{tab:full_ablation}
    \begin{tabular}{l|ccccc} 
        \toprule
        \textbf{Method} & \textbf{IQA} $\uparrow$ & \textbf{ASE} $\uparrow$ & \textbf{Sync-C} $\uparrow$ & \textbf{HKC} $\uparrow$ & \textbf{HKV} $\uparrow$ \\
        \midrule
        \multicolumn{6}{l}{\textit{Ablation on Agentic Reasoning}} \\
        \midrule
        Ours w/o Multi-Step Reasoning & 4.795 & 3.901 & 3.853 & 0.576 & 157.638 \\
        Ours w/o Analyzer             & 4.793 & 3.910 & 4.278 & 0.572 & 148.381 \\
        Ours w/o Reasoning (System 1 Only) & 4.784 & 3.885 & 3.507 & 0.544 & 122.376 \\
        \midrule
        \multicolumn{6}{l}{\textit{Ablation on Conditioning Modules}} \\
        \midrule
        Ours w/ Cross-Attention       & 4.745 & 3.856 & 3.263 & 0.558 & 116.317 \\
        Ours w/o MM-Warmup            & 4.752 & 3.866 & 3.993 & 0.549 & 164.080 \\
        Ours w/ Ref. Image            & 4.772 & 3.896 & 3.982 & 0.559 & 160.889 \\
        Ours w/o Ref. \& Pseudo Frame  & 4.682 & 3.878 & 4.141 & 0.564 & 160.986 \\
        \midrule
        \textbf{Ours (Full Model)}    & 4.790 & 3.901 & 4.087 & 0.571 & 168.912 \\
        \bottomrule 
    \end{tabular}
\end{table}
  
\begin{table}[t!]
    \caption{\textbf{Pairwise subjective ablation study and component comparison.} We report Lip-sync Inconsistency (LSI), Motion Unnaturalness (MU), Image Distortion (ID), and an overall Good/Same/Bad preference score (GSB). Lower is better for LSI, MU, and ID.}
    \label{tab:subjective_ablation_full}

    \begin{subtable}[t]{0.3\textwidth} \scriptsize
        \centering
        \caption{Ablation on agentic reasoning.}
        \label{tab:ablation_reasoning}
        \setlength{\tabcolsep}{1pt}
        \begin{tabular}{lcccc}
            \toprule
            Method & {LSI$\downarrow$} & {MU$\downarrow$} & { ID$\downarrow$} & { GSB$\uparrow$} \\
            \midrule
            Ours (w/o Reasoning) & 0.12 & 0.58 & 0.11 & $-$0.29 \\
            Ours (Full Model) & 0.12 & 0.37 & 0.04 & $+$0.29 \\
            \bottomrule
        \end{tabular}
    \end{subtable}
    \hfill
    \begin{subtable}[t]{0.3\textwidth} \scriptsize
        \centering
        \caption{Comparison of conditioning.}
        \label{tab:comparison_conditioning}
        \setlength{\tabcolsep}{1pt}
        \begin{tabular}{lcccc}
            \toprule
            Conditioning Method & { LSI$\downarrow$} & { MU$\downarrow$} & { ID$\downarrow$} & { GSB$\uparrow$} \\
            \midrule
            Previous Work~\cite{lin2025omnihuman} & 0.21 & 0.39 & 0.17 & $-$0.23 \\
            Ours (Proposed) & 0.03 & 0.25 & 0.07 & $+$0.23 \\
            \bottomrule
        \end{tabular}
    \end{subtable}
    \hfill
\begin{subtable}[t]{0.3\textwidth}
    \scriptsize
    \centering
    \caption{GSB Score Comparisons} 
    \label{tab:comparison_base}
    \setlength{\tabcolsep}{1pt} 
    \begin{tabular}{@{} l ccc} 
        \toprule
        GSB Comparisons & TA $\uparrow$ & Mot $\uparrow$ & VQ $\uparrow$ \\
        \midrule
        \rule[-1.8ex]{0pt}{5.2ex} 
        Ours vs. Base Model & -0.02 & +0.18 & +0.14 \\
        \bottomrule
    \end{tabular}
\end{subtable}
\end{table}
\par

\myparagraph{The Effectiveness of Proposed Conditioning Modules.}
We now ablate our core architectural designs, with results presented alongside the previous study in Table~\ref{tab:full_ablation} and~\ref{tab:subjective_ablation_full}. In these experiments, the Agentic Reasoning module remains fixed, providing identical inputs to all model variants. We test several key variations: using standard cross-attention for audio integration instead of our MM-Attention, removing the MM-Warmup strategy, conditioning on a reference image and omitting the pseudo-last-frame at inference.
As shown in Table~\ref{tab:full_ablation}, our full model again leads in most objective metrics, with its superior HKC and HKV scores highlighting enhanced motion dynamics. To further validate our approach, (b) of the Table~\ref{tab:subjective_ablation_full} presents a direct subjective comparison against OmniHuman-1~\cite{lin2025omnihuman}, a state-of-the-art method that utilizes a reference attention mechanism and standard cross-attention for audio injection instead of our proposed conditioning implementation. The results show that our method achieves a significant advantage not only in the overall GSB score but also across multiple fine-grained dimensions, including lip-sync accuracy, motion naturalness, and visual quality. This clearly demonstrates the effectiveness of our proposed conditioning technique, which in turn provides a robust foundation for executing the plans generated by the agentic reasoning module.

\par
In addition to the ablation studies, Table~\ref{tab:subjective_ablation_full} presents a direct comparison with the base model under text-only conditioning. For this evaluation, the audio component was disregarded to isolate the assessment of visual fidelity and character motion. We conducted a GSB pairwise comparison to analyze performance across three key areas: text alignment (TA), motion naturalness (Mot), and visual quality (VQ). The results reveal that our model successfully integrates multi-modal inputs while preserving a text-prompt-following capability on par with the pre-trained general model. Critically, our approach also demonstrates a significant lead in both motion naturalness and overall visual quality, as reflected by the Mot. and VQ metrics.

\subsection{Further Exploration on Applications}

\begin{table}[t!]
    \centering

    \setlength{\tabcolsep}{6pt} 

    \caption{\textbf{Comparison with existing methods on multi-person animation.} We report quantitative metrics and pairwise subjective evaluation results, including Driving Accuracy (DA), Lip-sync Inconsistency (LSI), Motion Unnaturalness (MU) and an overall user preference score derived from a Good/Same/Bad (GSB) evaluation.}
    \label{tab:sota_comparison_multi_final}
    \label{tab:sota_comparison_multi}

    \begin{tabular}{l|cccc|ccccc} 
        \toprule
        \multirow{2}{*}{\textbf{Method}} & \multicolumn{4}{c|}{\textbf{Subjective Evaluation}} & \multicolumn{5}{c}{\textbf{Quantitative Metrics}} \\
        \cmidrule(lr){2-5} \cmidrule(lr){6-10}
        & {\small DA$\uparrow$} & {\small LSI$\downarrow$}  & {\small MU$\downarrow$} & {\small GSB$\uparrow$} & {\small IQA$\uparrow$} & {\small ASE$\uparrow$} & {\small Sync-D$\downarrow$} & {\small HKC$\uparrow$} & {\small HKV$\uparrow$} \\
        \midrule
        InterActHuman       & {-}   & {-}   & {-}   & {-}      & 4.574 & 3.643 & 8.163 & 0.553 & 103.91 \\
        Ours w/o Reasoning  & 0.88 & 0.13 &  0.63 & -0.26   & 4.576 & 3.631 & 7.541 & 0.611 & 138.43 \\
        \midrule
        \textbf{Ours (Full Model)} & 0.94 & 0.04 & 0.12 & +0.26   & 4.529 & 3.653 & 6.904 & 0.614 & 158.36 \\
        \bottomrule
    \end{tabular}
\end{table}

\begin{figure}[t!]
  \centering
  \includegraphics[width=0.95\linewidth]{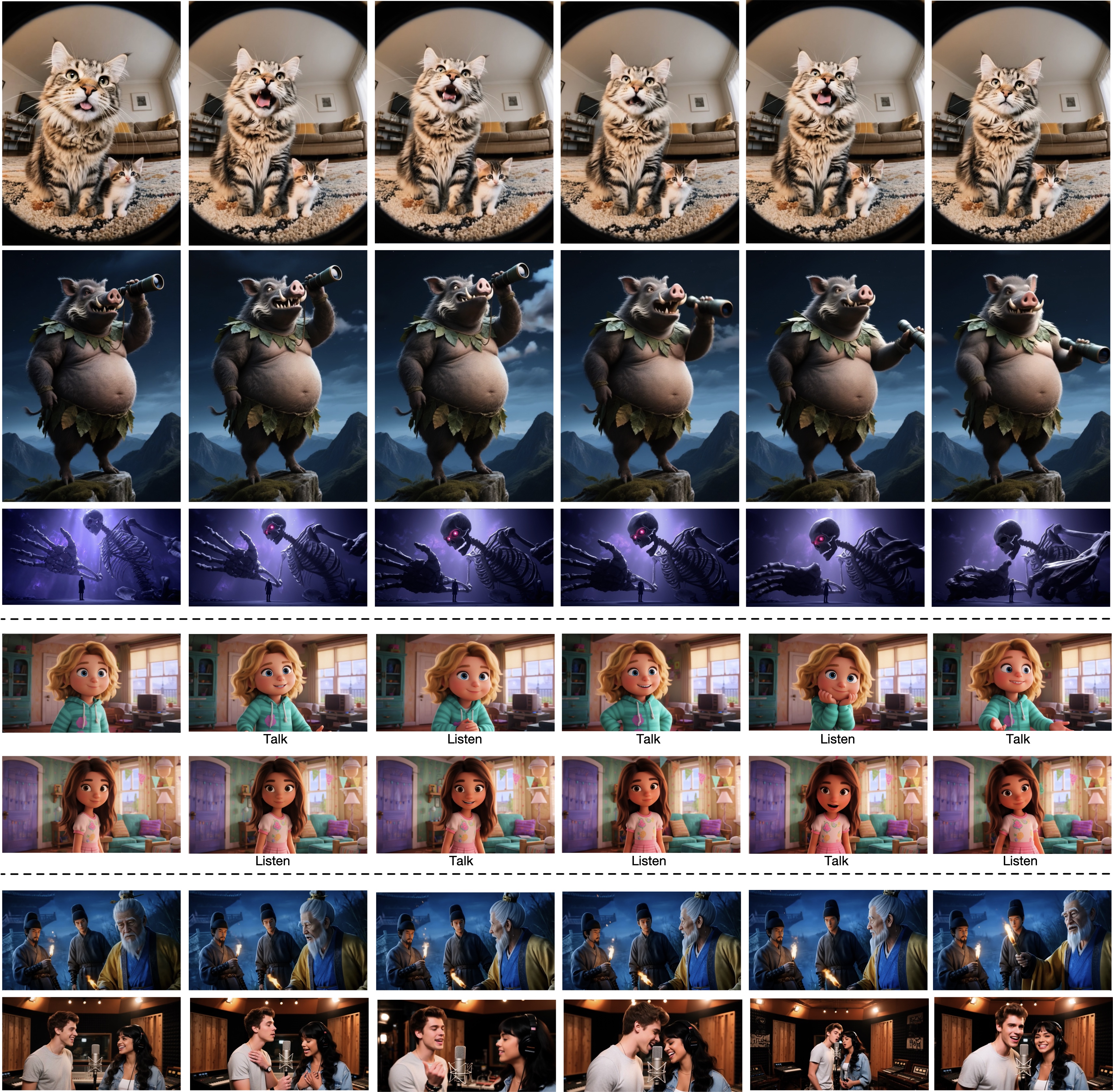}
  \caption{\textbf{Generalization and Multi-Person Results.} The top rows show the model's generalization across various non-human subjects. The fourth row presents a dialogue scenario, where characters correctly respond to conversational audio by switching between speaking and idle states. The bottom rows showcase performance in multi-person scenes, with coordinated behavior for both speakers and listeners.}
  \label{fig:diversity}
\end{figure}

\myparagraph{Applications on Diverse Inputs.} 
As depicted in Figure~\ref{fig:diversity}, we also explore the generalization capabilities of our model on non-human subjects, including anthropomorphic and animal characters. The results demonstrate remarkable robustness, which we attribute to our dual-system framework that effectively integrates high-level understanding with low-level synthesis.
Furthermore, the fourth row of the Figure~\ref{fig:diversity} highlights the model's capacity for conversational understanding, enabled by the agentic reasoning module. For the top and bottom images, we provide the same conversational dialogue but input the audio track corresponding to a different speaker for each. As shown, the characters seamlessly transition between speaking and idle states, correctly reflecting the turn-taking in the dialogue. This capability, when combined with efficient model acceleration techniques, showcases the potential of our framework for real-time, interactive conversational agent applications.

\myparagraph{Applications on Multi-person Scenarios.} 
  To enable multi-person animation, we extend our model with two modifications. First, we condition the synthesis on a speaker-specific mask that directs audio feature injection exclusively to the masked regions during the multimodal attention process. Following InterActHuman~\cite{wang2025interacthuman}, we employ a lightweight, plug-and-play predictor to dynamically generate these masks, ensuring robust speaker tracking through movement and occlusion without affecting the baseline single-person model. Second, we leverage our framework's inherent agent-based design by augmenting the Planner to also accept this mask to identify the active speaker. With the rest of the reasoning pipeline remaining identical, this simple extension enables the model to generate logically consistent and coordinated actions for all individuals in the scene. 
    \par
  As shown in Table~\ref{tab:sota_comparison_multi}, we present a quantitative and subjective comparison on our multi-person test set. Our full model, equipped with the agentic reasoning module, demonstrates significant improvements over two baselines that lack this capability: our model without agentic reasoning (ablation) and InterActHuman. Specifically, our method shows a clear advantage on metrics measuring gesture motion dynamics (HKC and HKV) and achieves better lip-sync accuracy. It is worth noting that we use Sync-D~\cite{syncnet} for evaluation, as the original Sync-C is effective for single-person lip-sync, and is less reliable for non-speaking individuals in multi-person scenarios. Furthermore, in pairwise subjective evaluations against the ablation model, our full model achieves higher driving accuracy (DA), defined as the ratio of correctly animated individuals (either speaking or silent) to the total number of people present. It also produces fewer instances of lip-sync inconsistency (LSI) and motion unnaturalness (MU). These subjective advantages are reflected in the overall win rate (GSB score), collectively validating the effectiveness of our proposed method.

\setlength{\tabcolsep}{3pt} 
\begin{table}[t]
    \centering
    
    \caption{\textbf{Quantitative comparison with audio-conditioned animation baselines.} 
    \textbf{(Left)} Portrait animation on the CelebV-HQ test set. 
    \textbf{(Right)} Full-body animation on the CyberHost test set.}
    \label{tab:quantitative_comparisons}

    \begin{minipage}[t]{0.41\textwidth}
        \centering
        \label{tab:portrait_comparison}
        \resizebox{\linewidth}{!}{
            \begin{tabular}{l|ccccc}
                \toprule
                \textbf{Method} & IQA $\uparrow$ & ASE $\uparrow$ & Sync-C $\uparrow$ & FID $\downarrow$ & FVD $\downarrow$ \\
                \midrule
                SadTalker         & 2.953 & 1.812 & 3.843 & 36.648 & 171.848 \\
                Hallo             & 3.505 & 2.262 & 4.130 & 35.961 & 53.992 \\
                EchoMimic         & 3.307 & 2.128 & 3.136 & 35.373 & 54.715 \\
                Loopy             & 3.780 & 2.492 & 4.849 & 33.204 & 49.153 \\
                Hallo-3           & 3.451 & 2.257 & 3.933 & 38.481 & 42.125 \\
                OmniHuman-1 & 3.875 & 2.656 & 5.199 & 31.435 & 46.393 \\
                \midrule
                Ours & \underline{3.817} & \textbf{2.663} & \underline{5.053} & \textbf{31.320} & \underline{45.771} \\
                \bottomrule
            \end{tabular}%
        }
    \end{minipage}\hfill
    \begin{minipage}[t]{0.58\textwidth}
        \centering
        \label{tab:full_body_comparison} 
        \resizebox{\linewidth}{!}{%
            \begin{tabular}{l|ccccccc}
                \toprule
                \textbf{Method} & IQA $\uparrow$ & ASE $\uparrow$ & Sync-C $\uparrow$ & FID $\downarrow$ & FVD $\downarrow$ & HKC $\uparrow$ & HKV $\uparrow$ \\
                \midrule
                Skyreel-A1        & 3.889 & 2.525 & 2.983 & 69.619 & 70.678 & 0.786 & 28.840 \\
                FantasyTalking       & 3.892 & 2.738 & 3.548 & 52.332 & 47.052 & 0.838 & 18.845 \\
                OmniAvatar        & 3.871 & 2.728 & 6.589 & 42.163 & 43.998 & 0.795 & 56.574 \\
                MultiTalk         & 3.822 & 2.681 & 6.868 & 37.308 & 32.783 & 0.817 & 62.753 \\
                OmniHuman-1 & 4.142 & 3.024 & 7.443 & 31.641 & 27.031 & 0.898 & 47.561 \\
                \midrule
                Ours & \textbf{4.144} & \textbf{3.030} & \underline{7.243} & \textbf{31.160} & \underline{27.642} & \underline{0.875} & \textbf{72.113} \\
                \bottomrule 
            \end{tabular}%
        }
    \end{minipage}
\end{table}
\setlength{\tabcolsep}{6pt}

\begin{figure}[t!]
\small
    \centering
    \caption{\textbf{Subjective User Preference Study.} We present results from two evaluation settings: (Left) a best-choice selection task comparing our method against academic baselines, and (Right) a GSB pairwise comparison against leading proprietary models.}
    \label{fig:combined_user_study}
    \begin{subfigure}[b]{0.35\textwidth}
        \centering
        \setlength{\tabcolsep}{4pt} 
        \begin{tabular}{l|c}
            \toprule
            \textbf{Method} & \textbf{Top-1 (\%)}  \\
            \midrule
            Skyreel-A1  & 5\% \\
            FantasyTalking  & 8\% \\
            OmniAvatar  & 14\%  \\
            MultiTalk  & 18\% \\
            OmniHuman-1  & 22\% \\
            \midrule 
            \textbf{Ours} & 33\% \\
            \bottomrule
        \end{tabular}
        \caption{Best-Choice Selection.}
        \label{tab:user_study_single_sub}
    \end{subfigure}
    \hfill
    \begin{subfigure}[b]{0.63\textwidth} 
        \centering
        \includegraphics[width=\linewidth]{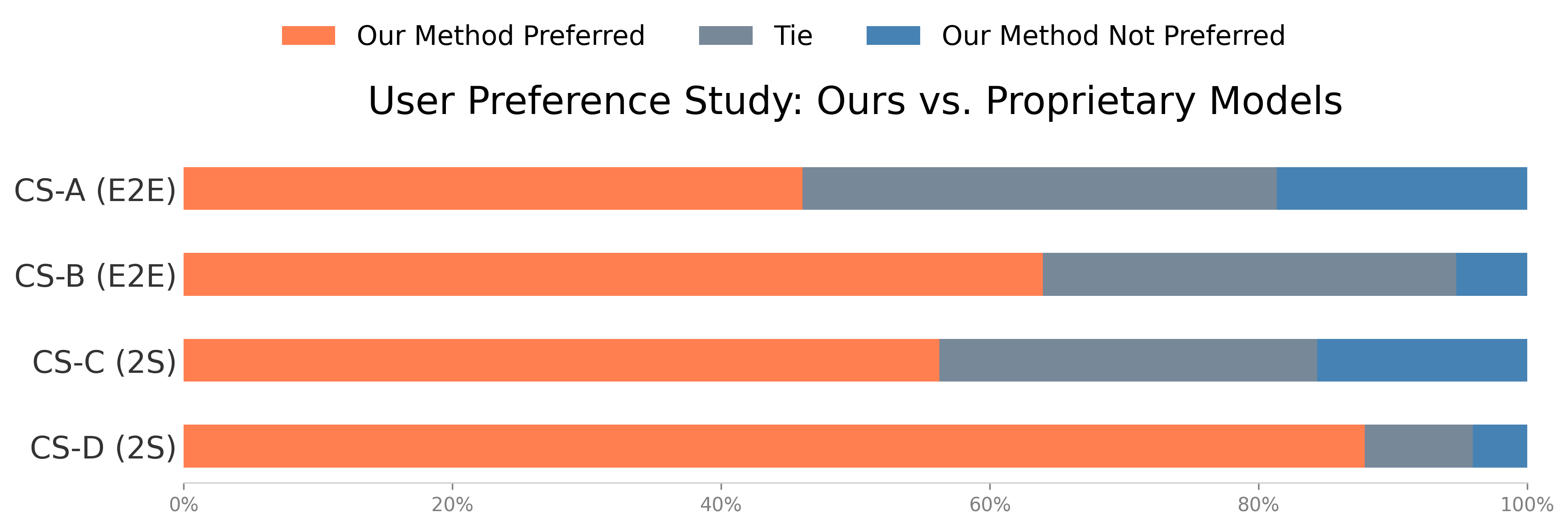} 
        \caption{GSB against leading proprietary models.}
        \label{fig:gsb_comparison_sub}
    \end{subfigure}

\end{figure}

\subsection{Comparison among Recent Methods}
\myparagraph{Comparisons with State-of-the-Art Methods.}
We conduct a comprehensive evaluation of our method against leading academic baselines across two distinct scenarios: portrait and full-body generation. For the portrait scenario, we compare our model against both specialized talking-head methods and state-of-the-art DiT-based approaches, including SadTalker~\cite{zhang2023sadtalker}, EchoMimic~\cite{chen2024echomimic}, Hallo~\cite{xu2024hallo}, Hallo3~\cite{hallo3}, Loopy~\cite{jiang2025loopy} and OmniHuman-1~\cite{lin2025omnihuman}, on the CelebV-HQ test set. For the more challenging task of full-body synthesis, our evaluation on the CyberHost test set includes a strong suite of recent DiT-based models: Skyreels-A1~\cite{fei2025skyreels}, FantasyTalking~\cite{wang2025fantasytalking}, OmniAvatar~\cite{gan2025omniavatar}, MultiTalk~\cite{kong2025multitalk} and OmniHuman-1~\cite{lin2025omnihuman}.
\par
The quantitative results in Table~\ref{tab:portrait_comparison} show our method consistently ranking in the top two across most metrics. In the portrait scenario, our model performs on par with the strong OmniHuman-1 baseline. We attribute this to the limited motion range in portrait videos, which challenges objective metrics in capturing subtle facial expressiveness. Our advantages become more pronounced in the full-body scenario. While leading in image quality and lip-sync, our model excels in generating dynamic, large-scale movements, evidenced by a high HKV score. Crucially, it achieves this without sacrificing local detail, maintaining a competitive HKC score. Taken together, these evaluation results demonstrate the clear advantages of our method over existing approaches.
\par
To provide a more holistic assessment of perceptual quality, we conducted user studies to supplement our quantitative metrics. First, we performed a user preference evaluation against the top academic baselines from our full-body comparison, with results shown in Figure~\ref{tab:user_study_single_sub}.
Additionally, we benchmarked our method against four leading proprietary models, which we categorized as either end-to-end (E2E) or two-stage (2S) systems (combining I2V and video dubbing). To comply with EULAs and avoid conflicts of interest, these models were anonymized as CS-A (E2E), CS-B (E2E), CS-C (2S), and CS-D (2S). The results of this comparison, shown in Figure~\ref{fig:gsb_comparison_sub}. These user studies demonstrate the superiority of our method, which is particularly evident in its handling of contextual coherence, a factor to which human users are highly sensitive yet one that objective metrics often fail to capture.
\begin{figure}[t!]
  \centering
  \includegraphics[width=\linewidth]{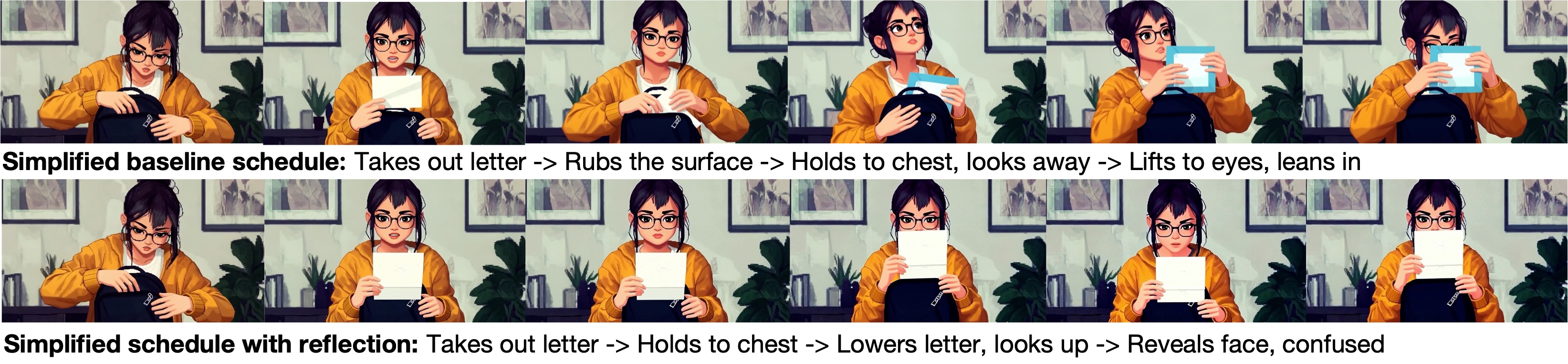}
  \caption{\textbf{Qualitative results of the reflection process.} Without reflection (first row), an ill-planned action ("Rubs the surface") causes object inconsistency. With reflection (second row), the model revises its plan to a more logical action, ensuring consistency.}
  \label{fig:vis1}
\end{figure}

\begin{figure}[t!]
  \centering
  \includegraphics[width=\linewidth]{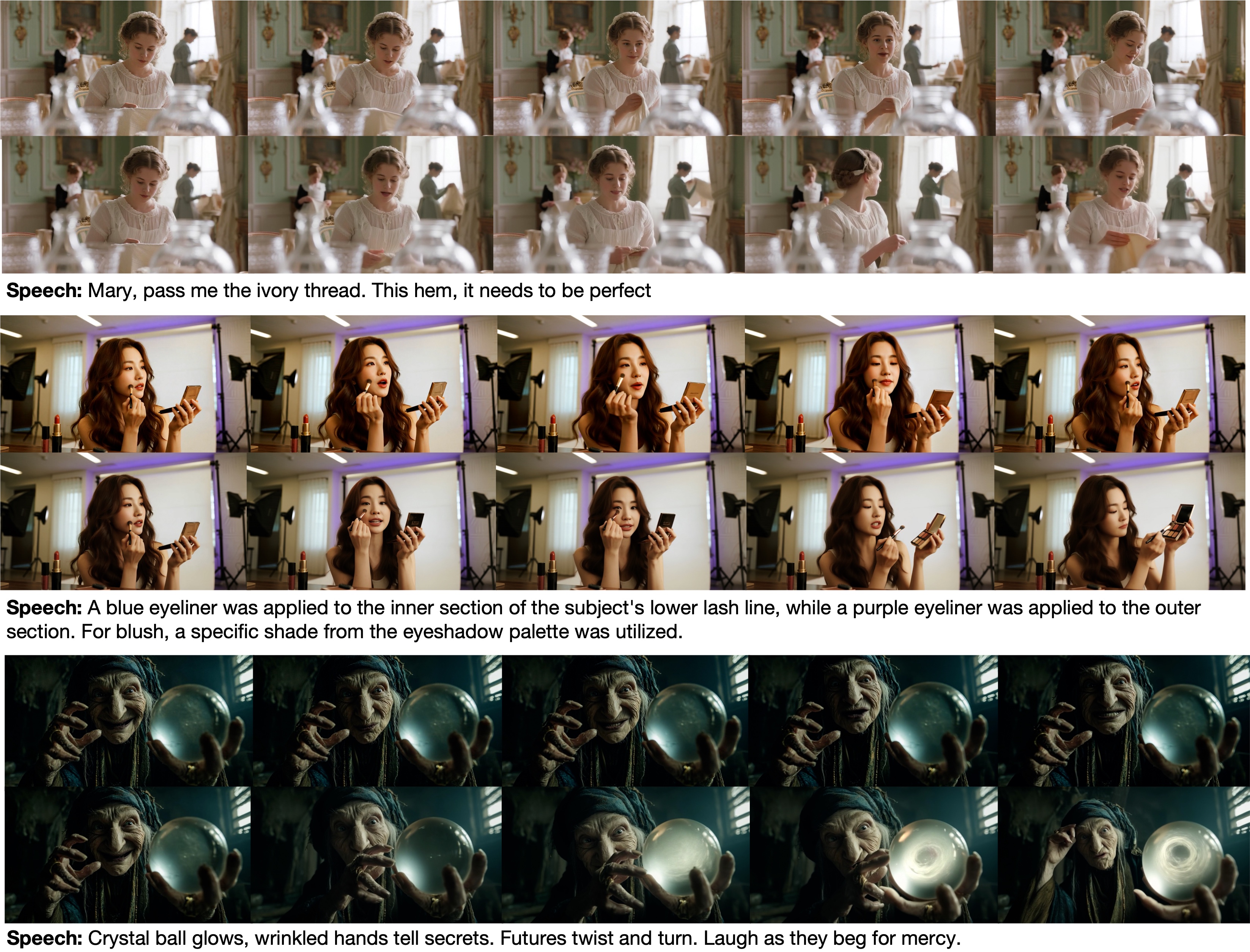}
  \caption{\textbf{Qualitative comparison of our model against OmniHuman-1~\cite{lin2025omnihuman}} For each pair of examples, our model (bottom row) generates actions with higher semantic consistency to the speech prompt than the baseline (top row). For example, our model correctly depicts a character applying makeup and a glowing crystal ball as described in the speech, actions which are absent in the baseline's results.}
  \label{fig:vis2}
\end{figure}

\subsection{Extended Experimental Results}

\myparagraph{Ablation Study on the Reasoning Process.}
We visualize the impact of our reflection process in Figure~\ref{fig:vis1}. This optional module is designed to correct errors from the initial action plan. Without reflection (top row), the model generates an action schedule in a single pass. This can lead to logical inconsistencies; for instance, after "Takes out letter," it generates "Rubs the surface," causing the letter to vanish and breaking semantic continuity.
In contrast, our model with reflection (bottom row) revises the plan after generating the first segment. It observes the outcome of "Takes out letter" and corrects subsequent actions to be relevant to the theme of letter-reading, ensuring logical progression and mitigating error accumulation. However, as this reflection process introduces additional inference overhead, it was disabled for the quantitative comparisons in this paper.
We also investigated injecting reasoning latents into the synthesis model. While this encouraged more nuanced facial expressions and subtle movements, it also suppressed large, dynamic actions. As this appeared to be more of an aesthetic trade-off than a clear improvement and was not a significant factor in user evaluations, we excluded it from our final model configuration.

\myparagraph{Visual Comparison with Baseline.}
In Figure~\ref{fig:vis2}, we present a visual comparison with OmniHuman-1, a strong baseline identified in our preceding experiments. The corresponding speech content for each video is provided. As can be seen, our method demonstrates significantly stronger semantic relevance and logical correlation between the audio and the generated actions. For instance, in the first video, the character turns her head when calling out "Mary"; in the second, she performs the specific actions of applying eyeliner and gesturing towards the eyeshadow palette as described; and in the third, the crystal ball glows and changes in response to the wizard's incantation. These high-level, context-aware results are difficult to capture with objective metrics and remain a challenge for existing methods. This qualitative evidence further substantiates the superiority of our approach. We encourage readers to view the examples on our project page for a more intuitive understanding of the correlation between generated motion, speech content and character intent. This alignment is a key aspect often overlooked in prior work.

\section{Conclusion}\label{sec:con}
In this work, we introduced a new paradigm for human video generation inspired by the dual-system theory of human cognition. We argued that existing methods primarily simulate reactive "System 1" thinking, failing to align motion with high-level intent. We proposed OmniHuman-1.5, a framework that additionally models deliberative "System 2" processes through two key innovations: an MLLM-based agent for semantic planning and a specialized MMDiT architecture with a novel pseudo last frame strategy to fuse multimodal signals.
Experiments show our approach generates more expressive and logically consistent results, which users significantly preferred for their naturalness and plausibility. By demonstrating this framework's effectiveness, even extending it to multi-person scenarios, we believe simulating cognitive agency offers a new perspective for creating the next generation of lifelike digital humans.

\section{Broader Impact}
Our core contribution in this work is to introduce a novel paradigm for video avatar generation. By simulating a dual-system cognitive framework, our model achieves a new level of expressive capability and logical coherence in motion, moving beyond the limitations of single-process generation. While this advancement opens up exciting possibilities for creative applications like AI-driven film production and music videos, we are acutely aware of the potential for misuse associated with highly realistic avatar technologies.
To address these ethical concerns, we advocate for a robust framework of responsible deployment. Although current results may still bear subtle artifacts of AI generation, which can serve as a minor deterrent, proactive safeguards are essential. We strongly recommend the following measures: (1) applying prominent, visible watermarks to all generated content to clearly label it as AI-generated; (2) implementing filtering algorithms to reject inappropriate or malicious input prompts and to review output content; and (3) embedding traceable, invisible watermarks to ensure accountability and aid in source identification if misuse occurs. By integrating these safety protocols, we can help ensure that our technology fosters creativity while minimizing the risks of malicious applications such as fraud or disinformation.

\section{Acknowledgment}
We thank Ashley Liang, Bei Li, Xu Song, Jie Ling, Wei Han and Xi Yang for their help with the evaluation. We also thank the ByteDance Seed Video Generation Team for their valuable discussions regarding model training.

\clearpage

\bibliographystyle{plainnat}
\bibliography{main}


\end{document}